\documentclass[runningheads,orivec]{llncs}

\usepackage{adjustbox}
\usepackage{amsmath}
\usepackage{amssymb}
\usepackage{array}
\usepackage{booktabs}
\usepackage{caption}
\usepackage{color}
\usepackage{enumitem}
\usepackage{epstopdf}
\usepackage{float}
\usepackage[T1]{fontenc}
\usepackage{framed}
\usepackage{graphicx}
\usepackage{hyperref}
\usepackage{cleveref} 
\usepackage{import}
\usepackage[utf8]{inputenc}
\usepackage{lineno}
\usepackage{listings}
\usepackage{lscape}
\usepackage{microtype}
\usepackage{multicol}
\usepackage{multirow}
\usepackage{newtxtext}
\usepackage{pifont}  
\usepackage{pstricks}
\usepackage{rotating}
\usepackage{siunitx}
\usepackage{soul}
\usepackage{subcaption}
\usepackage{subfiles}
\usepackage{svg}
\usepackage{tabularx}
\usepackage{tcolorbox}
\usepackage{umoline}
\usepackage{url}
\usepackage{verbatim}
\usepackage{wrapfig}
\usepackage{xcolor}
\usepackage{xstring}
\usepackage{xspace}

\newcommand{\xmark}{\ding{55}}

\newcommand{\pmrnumber}{nine }

\newcolumntype{R}{>{\raggedleft\arraybackslash}X}
\newcolumntype{C}{>{\centering\arraybackslash}X}

\begin{document}

\title{What is the Best Process Model Representation? \\ A Comparative Analysis for Process Modeling with Large Language Models}

\titlerunning{What is the Best Process Model Representation ?}

\author{Alexis Brissard\inst{1,2,3}\orcidID{0009-0005-1710-0133} 
\and
Frédéric Cuppens\inst{1,2}\orcidID{0000-0003-1124-2200} 
\and
Amal Zouaq\inst{1,3}\orcidID{0000-0002-4791-0752}}
\authorrunning{A. Brissard et al.}
%
\institute{Polytechnique Montréal
\and
LabCys
\and
LAMA-WeST Lab\\
\email{\{first name.last name\}@polymtl.ca}
}

\maketitle              

\begin{abstract}
Large Language Models (LLMs) are increasingly applied for Process Modeling (PMo) tasks such as Process Model Generation (PMG). To support these tasks, researchers have introduced a variety of Process Model Representations (PMRs) that serve as model abstractions or generation targets. However, these PMRs differ widely in structure, complexity, and usability, and have never been systematically compared. Moreover, recent PMG approaches rely on distinct evaluation strategies and generation techniques, making comparison difficult.

This paper presents the first empirical study that evaluates multiple PMRs in the context of PMo with LLMs. We introduce the PMo Dataset, a new dataset containing 55 process descriptions paired with models in nine different PMRs. We evaluate PMRs along two dimensions: suitability for LLM-based PMo and performance on PMG. \textit{Mermaid} achieves the highest overall score across six PMo criteria, whereas \textit{BPMN text} delivers the best PMG results in terms of process element similarity.

\keywords{Business Process Modeling \and Large Language Models \and Process Model Generation \and Process Model Representation}
\end{abstract}

\section{Introduction}\label{sec:introduction}

\subsection{Background}

One of the most promising applications of LLMs in BPM is Process Modeling (PMo), where the goal is to assist or automate the construction of process models from textual process descriptions~\cite{grohs_large_2023}.
A central subtask in PMo is Process Model Generation (PMG), which involves generating a process model from a natural language input. To support this generation, researchers have introduced a variety of Process Model Representations (PMRs)—textual abstractions used to represent the output process model. 

\begin{figure*}[tbp]
\centering
\includegraphics[width=1.0\textwidth]{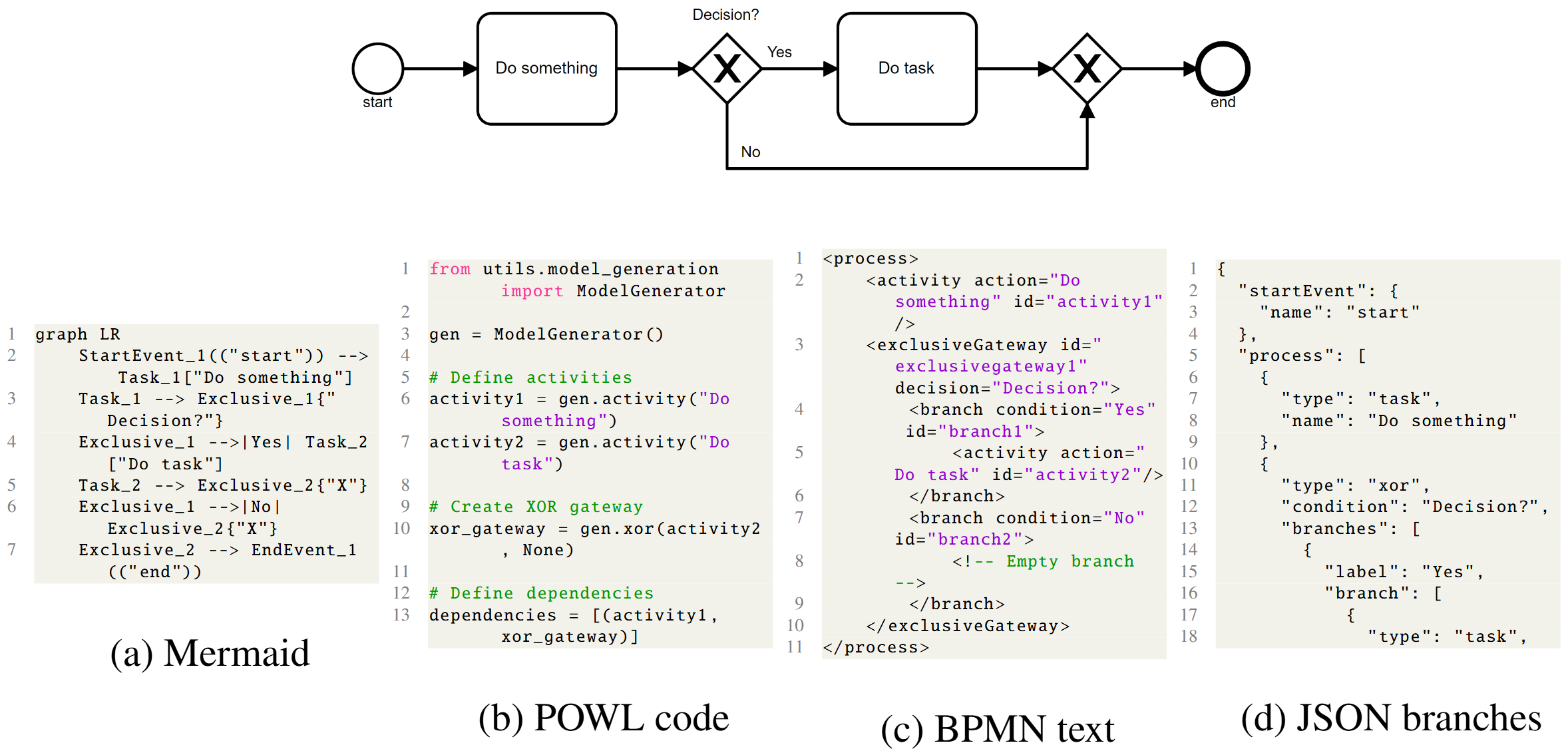} 
\caption{Four Process Model Representations (PMRs) of the same process model.}
\label{fig:pmr_overview}
\end{figure*}

The initial motivation for using PMRs stemmed from the limitations of LLMs, which were unable to directly generate standard BPMN models due to constraints in context length and suboptimal LLM performance~\cite{grohs_large_2023,klievtsova_conversational_2024}. To address these challenges, alternative representations such as simplified BPMN or existing diagram notations were introduced. Since then, more advanced PMG approaches have emerged, proposing new techniques such as constrained generation or multi-agents framework, often taking advantage of their own custom PMRs~\cite{kourani_promoai_2024,lin_mao_2024,voelter_leveraging_2024,kopke_efficient_2024}. A subset of these PMRs are presented in \Cref{fig:pmr_overview}.

Despite this progress, the evaluation of the generated process models remains a major challenge~\cite{fettke_evaluating_2025}. Each approach introduces its own evaluation methods, metrics, and benchmarks, preventing meaningful comparisons between them. Furthermore, the underlying PMRs differ significantly between approaches and have never been subject to a systematic comparative analysis. This fragmentation makes it difficult to assess which PMRs are most suitable for PMG and more broadly for PMo tasks.

\subsection{Research Questions and Contributions}

This work systematically investigates PMRs by assessing their respective strengths and limitations through conceptual and empirical evaluation. 
To clarify the current landscape and guide future research in the field, we address the following research questions:

\textbf{RQ1}: What are the key characteristics and features that distinguish different PMRs?

\textbf{RQ2}: Which PMRs are most suitable for PMo with LLMs?

\textbf{RQ3}: Which PMRs lead to the best performance on PMG?

To answer these research questions, we make the following contributions.
First, we introduce the PMo Dataset, a new dataset comprising 55 process descriptions paired with process models in \pmrnumber distinct PMRs. 
Second, we perform the first comprehensive review and intrinsic comparison of PMRs in the context of PMo with LLMs, analyzing their structural properties and representational characteristics using both qualitative and quantitative methods.
Finally, we evaluate the effectiveness of each PMR as a target for PMG under standard prompting with LLMs, identifying which representations facilitate more accurate and faithful model generation.

The code of this project, containing all our data, prompts and results can be found at 
\href{https://github.com/Lama-West/Process_Model_Representations}{\texttt{https://github.com/Lama-West/Process\_Model\_Representations}.}

\section{Related Work}\label{sec:related_work}

\subsubsection{PMo with LLMs.}\label{sec:pmo_llms}

LLMs have recently started to be used for Process Modeling (PMo) and show promising results\cite{kourani_evaluating_2024,fettke_evaluating_2025,klievtsova_conversational_2024}.
\cite{klievtsova_conversational_2024} have framed their approach as \textit{Conversational PMo}, we use the broader formulations \textit{PMo with LLMs} or \textit{LLM-based PMo}.

\begin{table}[t]
\centering
\caption{Main PMG approaches from the literature.}
\label{tab:main_pmg_approaches}
\begin{tabular}{p{2.7cm}p{2.2cm}p{1.2cm}p{2.7cm}p{2.2cm}}
\toprule
\textbf{PMG approach} & 
\textbf{PMR} & 
\textbf{LLM} & 
\textbf{Evaluation set} & 
\begin{tabular}[c]{@{}l@{}}\textbf{Evaluation} \\ \textbf{method}\end{tabular} \\ 
\midrule
ConverMod~\cite{klievtsova_conversational_2024}        & \begin{tabular}[c]{@{}l@{}}Graphviz +\\ Mermaid\end{tabular}               & GPT-4                              & \begin{tabular}[c]{@{}l@{}}Mangler + PET-7\end{tabular} & \begin{tabular}[c]{@{}l@{}}Text similarity +\\ Qualitative\end{tabular}                \\
PromoAI~\cite{kourani_promoai_2024}                    & POWL code                        & GPT-4                              & 2 descriptions                                            & Qualitative                                    \\
MAO~\cite{lin_mao_2024}                                & BPMN text                        & GPT-4                              & BPMN for Research                                         & BPMNDiff Viz                                   \\
Multimodal~\cite{voelter_leveraging_2024}              & PME                              & GPT-4V                             & SAP-SAM                                                   & PME similarity                                 \\
BPMN-chatbot~\cite{kopke_efficient_2024}               & JSON branches                    & GPT-4o                             & PET-7                                                     & Qualitative                          \\
\bottomrule
\end{tabular}
\end{table}

\subsubsection{Main PMG approaches.}\label{sec:pmg_approaches}

One of the most complete PMG studies is ConverMod~\cite{klievtsova_conversational_2024}, investigating the use of Graphviz and Mermaid as PMRs. They evaluate their approach on a subset of 7 documents from the PET dataset~\cite{bellan_pet_2022} for which they create business models (hereafter referred to as PET-7). Their main evaluation metric is the Jaccard index between the generated and gold sequence flows.

\cite{kourani_promoai_2024,kourani_process_2024} introduce ProMoAI, a unique approach where Python code is generated and executed to obtain POWL process models. Designated as \textit{POWL code}, this PMR requires extended prompts and multiple error correction rounds, which limits its practicality.

\cite{kopke_efficient_2024} proposes BPMN-chatbot, an enhancement of the ProMoAI framework that introduces a more structured and practical PMR. Labeled in this study as \textit{JSON branches}, this PMR encodes the process branching structure in JSON format and leverages OpenAI's function calling to enforce schema adherence during generation. Using PET-7 as in Convermod, the authors conduct a qualitative evaluation and report improved performance over both previous approaches.

\cite{lin_mao_2024} uses a multi agent framework combined with a custom PMR (BPMN text) to improve the generation process. They evaluate their results on the BPMN for research dataset~\cite{camunda_bpmn_2015} using BPMNDiffViz~\cite{ivanov_bpmndiffviz_2015}, a tool to calculate the structural similarity between 2 processes. According to these metrics, their method outperform ProMoAI.

\cite{voelter_leveraging_2024} evaluates the multimodal capabilities of LLMs by generating process models from process description and diagrams pairs. They introduce a PMR based on lists of process model elements (PME), used for both generation and evaluation. 

Table~\ref{tab:main_pmg_approaches} summarizes PMG approaches presented in this section. Despite recent progress, existing studies face several limitations: reliance on closed-source LLMs limits reproducibility, evaluation sets remain small, and performance is often assessed using non-standardized metrics. While some studies have begun to compare LLMs~\cite{kourani_evaluating_2024}, the field still lacks a systematic and reproducible evaluation of PMG pipelines. In response, our study starts to address these gaps through a unified benchmark and a comparative framework, introduced in the following section.

\section{Methodology}\label{sec:methodology}

\subsection{Initial Definitions}\label{sec:definitions}

To ensure clarity throughout the paper, we distinguish between the following concepts:

\textbf{Process model}: An abstract model of a business process, independent of any specific formalism or notation. For brevity, we occasionally refer to it simply as a \textit{model}.
    
\textbf{Process Model Representation (PMR)}: A notation used to express a process model (e.g., BPMN, POWL, or Mermaid). A PMR defines the syntax and semantics by which a process model is described.
    
\textbf{PMR model}: A process model represented using a PMR (e.g., a BPMN model, a Mermaid model).

\subsection{Process Model Representations}\label{sec:pmrs}

\begin{table}[t]
\caption{PMR main features. PMRs are ordered by recency.}
\label{tab:pmr_main_features}
\begin{tabularx}{\textwidth}{@{}llCCCCC@{}}
\toprule
\textbf{PMR} & \textbf{Language} & \textbf{\begin{tabular}[c]{@{}r@{}}Graph\\ based\end{tabular}} & \textbf{\begin{tabular}[c]{@{}c@{}}Branch\\ based\end{tabular}} & \textbf{Executable} & \textbf{Visualizable} & \textbf{\begin{tabular}[c]{@{}c@{}}Generation\\ schema\end{tabular}} \\
\midrule
BPMN                             & XML                                   & $\checkmark$         & \xmark                & $\checkmark$        & $\checkmark$          & \xmark                     \\
BPMN process                     & XML                                   & $\checkmark$         & \xmark                & $\checkmark$        & \xmark                & \xmark                     \\
Graphviz                         & Graphviz                              & $\checkmark$         & \xmark                & \xmark              & $\checkmark$          & \xmark                     \\
Mermaid                          & Mermaid                               & $\checkmark$         & \xmark                & \xmark              & $\checkmark$          & \xmark                     \\
PME                              & JSON                                  & $\checkmark$         & \xmark                & \xmark              & \xmark                & $\checkmark$               \\
Simplified XML                   & XML                                   & $\checkmark$         & \xmark                & \xmark              & \xmark                & \xmark                     \\
POWL code                        & Python                                & \xmark               & \xmark                & $\checkmark$        & \xmark                & \xmark                     \\
BPMN text                        & XML                                   & \xmark               & $\checkmark$          & \xmark              & \xmark                & \xmark                     \\
JSON branches                    & JSON                                  & \xmark               & $\checkmark$          & \xmark              & \xmark                & $\checkmark$               \\ 
\bottomrule
\end{tabularx}
\end{table}

We choose to evaluate \pmrnumber PMRs, presented in \Cref{tab:pmr_main_features}.
Our selection is guided by two criteria. First, we include PMRs that are being used in existing PMG approaches, ensuring relevance to the current state of the art. Second, we ensure diversity in representation style, covering a range of structures and languages.

Our analysis focuses on a specific set of element types commonly found in process models. 
These include tasks, events, gateways (with optional labels referred to as \textit{decisions}) and sequence flows (with optional labels referred to as \textit{conditions}). Furthermore, we consider swimlanes, which feature lanes nested within pools, and message flows.
We do not include data objects, textual annotations or other special elements due to their limited prevalence in the PMo Dataset.

Hereafter, we describe the main PMRs used in this study.
\textbf{BPMN} is selected as our baseline given its status as a widely adopted standard. 
\textbf{BPMN process} 
is a BPMN model that only defines the process and does not include the BPMN diagram definition. Comparing BPMN and BPMN process allows us to directly measure the impact of specifying graphical elements.
\textbf{Graphviz} 
is a graphical representation, meaning it can be directly visualized using the DOT engine. 
\textbf{Mermaid} 
is another graphical representation made for diagrams. 
\textbf{Process Model Elements} 
(PME)~\cite{voelter_leveraging_2024} consists of a JSON object containing 6 lists of elements: tasks, events, gateways, swimlanes, sequence flows and message flows. 
This flat structure enables easy counting of elements and comparison between models. 
\textbf{Simplified XML} 
is a simplification of XML originally proposed in \cite{kourani_leveraging_2024}. Similarly to BPMN process, graphical and other non-essential elements are omitted. The XML structure is lightened, resulting in a more compact PMR.
\textbf{POWL code}, introduced by \cite{kourani_promoai_2024}, is Python code that can be executed to create a POWL model, itself translatable to BPMN.
The main advantages of using code as a PMR is to leverage LLM's familiarity with coding tasks while enabling detailed feedback via execution of said code. POWL code lacks support for conditions, event labels, intermediate events, swimlanes and message flows, making it the least expressive PMR.
\textbf{JSON branches} 
was developed by \cite{kopke_efficient_2024} specifically to reduce token count, be compact and provide schema following abilities for LLM generation. 
It only handles simple backward flows (represented as looping gateways) and cannot include multiple start and end events, message flows and swimlanes.
\textbf{BPMN text}~\cite{lin_mao_2024} is similar to JSON branches, but in XML format. It has the same limitations except that it supports role and object annotations in tasks.

\subsection{PMo Dataset}\label{sec:pmo_dataset}

To enable PMR comparison across a wide range of process models, we build the PMo dataset. It contains 55 process descriptions paired with process models represented in all PMRs from \Cref{sec:pmrs}.
The PMo dataset is constructed from five existing sources: 24 pairs from the Mangler dataset~\cite{mangler_textual_2023}, 20 from the PMo Benchmark~\cite{kourani_evaluating_2024}, 6 from the PET-7 dataset~\cite{klievtsova_conversational_2024}\footnote{One pair is removed because its description is already included in the PMo Benchmark.}, 4 from the BPMN for research dataset~\cite{camunda_bpmn_2015}, and 1 from the CCC19 dataset~\cite{munoz-gama_conformance_2019}.\footnote{The PMo Dataset is accessible at this url: \href{https://doi.org/10.5281/zenodo.15857588}{\texttt{https://doi.org/10.5281/zenodo.15857588}}}

We chose these data sources because process models are handcrafted, or at least validated by domain experts. This explains why we did not include the much larger MaD dataset~\cite{li_mad_2023}, which has been criticized for lack of variability of its models and their descriptions. Similarly, even though the descriptions are human-authored, the dataset in \cite{apaydin_local_2025} also lacks diversity due to automatic process model generation (e.g., maximum 9 activities).

The PMo Dataset underwent extensive preprocessing to ensure optimal usability and applicability. For process descriptions, this involves cleaning special characters, correcting punctuation and spacing, sentence splitting, and removing irrelevant information such as modeling instructions. The ground truth BPMN models are also refined by sanitizing label texts, improving diagram layouts and positioning decisions and conditions optimally.
The Mangler dataset received special attention due to having multiple models per description, each graded from 0 to 5. The model closest to the textual description (to our own judgment) is chosen among those with the best grade, with a preference for models including only common elements (e.g., no data objects or other special elements).

Following the preprocessing phase, we automatically convert all BPMN models into all the other representations to obtain our ground truth. We develop converters for each PMR and validate our conversions by transforming models from BPMN to PMRs and back.
Some process models contain information that are not supported by every PMR. In this case, we ignore the additional information (e.g., conditions are not included in POWL code). In cases where the model cannot be represented by the PMR without significant loss of information, it is left out (e.g., models including swimlanes are not converted to JSON branches). For branching PMRs, two thirds of the BPMN models cannot be converted.

\subsection{PMR for PMo with LLMs}\label{sec:intrinsic_comparison}

\subsubsection{Requirements.}\label{sec:requirements}

To guide our evaluation, we define six key requirements for an ideal PMR in the context of PMo with LLMs.
A suitable PMR should be compact in terms of token count to fit within LLM context limits, reduce costs, generation time and energy consumption~\cite{kopke_efficient_2024}.
Strong expressiveness is essential, to ensure that a wide range of models can be represented. This is mainly measured by element coverage, i.e. how many elements from a model are representable by the PMR.
The PMR should be human readable to support interpretability. More specifically, a non expert should be able to understand the general purpose and structure of the process by just looking at the PMR model itself~\cite{klievtsova_conversational_2024}. Length in lines, words and characters are key metrics to consider for this requirement, as a longer model is often harder to understand.
Beyond pure text, a PMR should should also support direct graphical visualization, ideally without the need for additional conversions.
The representation must be easily usable—parsable, editable, and generable by LLMs. This includes support for additional tooling such as JSON schema.
Finally, it should be possible to extend it to fit specific modeling needs or add support for other process model elements.

In summary, our six requirements are Token compactness, Expressivity, Human readability, Visualization capabilities, Usability and Extensibility. These requirements are specific to the use case of PMo with LLMs and are designed to support the whole range of associated tasks (e.g., question answering, interactive modification). They represent our attempt to differentiate between PMRs using the PMo Dataset. Other criteria could also be included such as compositionality, but it would require a more advanced evaluation framework and dataset.

\subsubsection{Evaluation.}\label{sec:instrinsic_evaluation}

We perform a comparison of PMRs across six dimensions that reflect the previously defined requirements.
This evaluation combines both quantitative and qualitative metrics. For quantitative analysis, we measure the length of each representation in terms of lines, tokens, words, and characters to assess their compactness and efficiency. We also compute the mean element coverage over the PMo Dataset by dividing the number of process elements representable in the PMR by the total number of ground truth elements. For qualitative analysis, we evaluate human readability, visualization capabilities, usability and extensibility.

\subsection{PMR for PMG}\label{sec:pmr_for_pmg}

We aim to measure the performance of each PMR for the specific task of PMG. 
To ensure a fair comparison, we perform all experiments under a standardized setting. Specifically, we restrict process elements to the following set: standard tasks, start and end events, exclusive and parallel gateways, and sequence flows. 
This simplified setting allows us to compare PMRs independently of their element coverage, while also limiting prompts length and complexity. Indeed, it would not be fair to compare a complex model generated in a PMR that supports swimlanes to a simpler model in a less expressive PMR. 

\subsubsection{Prompting.}\label{sec:prompting}

We design a PMG prompt that instructs the LLM to generate a process model-represented in a specific PMR based on a given process description. 
To ensure a fair evaluation across PMRs, we construct prompts with consistent size and structure. The LLM receives the same task description and general modeling instructions regardless of the PMR. For each representation, we provide specific formatting guidelines which include Markdown subsections and an example illustrating the expected output format. 

\subsubsection{LLM Choice and Generation Parameters.}\label{sec:llm_choice}

We perform our experiments with LLaMA-3.3-70b, which is a recent, open-source model with state of the art performance for its size. 
Our choice to use an open-source LLM is motivated by transparency, adaptability, and community support, which facilitate reproducibility and further research. While it could be interesting to evaluate additional LLMs, this is out of the scope of this paper.
All experiments are performed using Google Vertex AI API.
Generation parameters are set as follows: Top-K is left as default (i.e., -1), Top-P is 0.95 and Temperature is 0.2.

\subsubsection{Evaluation.} \label{sec:pmr_for_pmg_evaluation}

We evaluate the generated PMR models through comparison with the ground-truth PMR models along two dimensions: element counts, and PME similarity.
Element counts capture the number of generated process elements for each type.
To also take in account the semantic content of elements (i.e., task labels, decisions, conditions), we follow the evaluation methodology proposed by \cite{voelter_leveraging_2024}. It consists of 2 steps: semantic matching and set similarity. Element pairs with semantic similarity higher than an experimentally defined threshold (0.7) are made identical using a 1-to-1 matching. To calculate this semantic similarity, we produce embeddings for each element using a sentence transformer trained for sentence similarity (\texttt{stsb-mpnet-base-v2}) and calculate their cosine similarity.
Once the elements are semantically matched, the Dice–Sørensen coefficient is computed to assess similarity between the resulting sets, balancing precision and recall in a single score.

\section{Results}\label{sec:results}

\subsection{PMR for PMo with LLMs}

\subsubsection{Length comparison.}

\Cref{tab:length_diff_bpmn} presents the length differences between ground-truth PMR models and our baseline, BPMN models. All PMRs lead to a significant reduction across all length metrics, successfully achieving their initial purpose. However, we observe substantial variation between PMRs: Mermaid is the most compact, achieving over 90\% reduction in lines, tokens, words, and characters compared to BPMN. Looking at BPMN process, we can see that removing the BPMN diagram definition alone leads to a 70\% decrease in token count. This highlights the importance of excluding non-essential elements such as graphical details.

POWL code, BPMN text, and JSON branches are reported in a separate section of the table as they do not cover all processes of the PMo Dataset, due to element coverage limitations. This explains the slight discrepancies between absolute and relative difference in the results. Nonetheless, they also achieve significant length reductions, comparable to Mermaid.

\begin{table}[t]
\centering
\caption{Mean length difference of ground-truth PMR models compared to BPMN models.}
\label{tab:length_diff_bpmn}

\begin{tabularx}{\textwidth}{@{}lRRRRRRRR@{}}
\toprule
\multicolumn{1}{l}{\multirow{2}{*}{\textbf{PMR}}} & \multicolumn{2}{r}{\textbf{Line Count}}                             & \multicolumn{2}{r}{\textbf{Token Count}}                            & \multicolumn{2}{r}{\textbf{Word Count}}                             & \multicolumn{2}{r}{\textbf{Char Count}}                        \\
\multicolumn{1}{c}{}                              & \multicolumn{1}{r}{\textbf{Rel}} & \multicolumn{1}{r}{\textbf{Abs}} & \multicolumn{1}{r}{\textbf{Rel}} & \multicolumn{1}{r}{\textbf{Abs}} & \multicolumn{1}{r}{\textbf{Rel}} & \multicolumn{1}{r}{\textbf{Abs}} & \multicolumn{1}{r}{\textbf{Rel}} & \multicolumn{1}{r}{\textbf{Abs}} \\ 
\midrule
BPMN (Baseline)                                   & \multicolumn{2}{r}{384}                                             & \multicolumn{2}{r}{4231}                                            & \multicolumn{2}{r}{6531}                                            & \multicolumn{2}{r}{19048}                                           \\
\midrule
BPMN process                                      & -64\%                            & -248                             & -70\%                            & -2984                            & -69\%                            & -4539                            & -63\%                            & -11947                           \\
Graphviz                                          & -87\%                            & -336                             & -88\%                            & -3734                            & -90\%                            & -5888                            & -88\%                            & -16803                           \\
Mermaid                                  & \textbf{-91\%}                   & \textbf{-350}                    & \textbf{-93\%}                   & \textbf{-3934}                   & \textbf{-93\%}                   & \textbf{-6095}                   & \textbf{-92\%}                   & \textbf{-17524}                  \\
PME                            & -27\%                            & -109                             & -65\%                            & -2783                            & -73\%                            & -4768                            & -62\%                            & -11799                           \\
Simplified XML                                    & -60\%                            & -228                             & -83\%                            & -3502                            & -79\%                            & -5111                            & -70\%                            & -13207                           \\
\midrule
POWL code                                         & -87\%                            & -280                             & -93\%                            & -3268                            & -94\%                            & -5107                            & -91\%                            & -14471                           \\
BPMN text                                         & -89\%                            & -284                             & -89\%                            & -3119                            & -93\%                            & -5032                            & -88\%                            & -13900                           \\
JSON branches                                     & -67\%                            & -212                             & -88\%                            & -3078                            & -92\%                            & -4973                            & -80\%                            & -12215                           \\ 
\bottomrule
\end{tabularx}
\end{table}

\subsubsection{Evaluation summary.}

\Cref{tab:pmr_for_pmo_summary} summarizes our evaluation of PMRs across the six requirements defined in \Cref{sec:requirements}. We graded each requirement from 1 to 5 based on specific quantitative and qualitative metrics detailed in the same section.

Token compactness reflects the results from \Cref{tab:length_diff_bpmn}, with Mermaid and POWL code achieving the highest scores.

For expressiveness, we report the element coverage calculated on the PMo Dataset models. Branching PMRs such as POWL code and BPMN text score lower due to their limited element support (see \Cref{sec:pmrs}).

Human readability scores are based on length results (\Cref{tab:length_diff_bpmn}) and ease of understanding of the process solely from the PMR model. Branching PMRs perform best in this regard due to their concise syntax and clearer structure. We also take in account the user study from \cite{klievtsova_conversational_2024}, where participants preferred Mermaid to Graphviz representation.

BPMN, Mermaid and Graphviz have the best visualization capabilities as they can be directly represented graphically. BPMN process and POWL code can be visualized via standard libraries such as pm4py, while the remaining PMRs need to be converted. The PMRs that are not graph-based are harder to convert, hence their reduced grade.

Usability scores are based on 3 aspects: parsability, editability and support for external tooling. 
Parsability is highest for JSON and XML-based formats, which benefit from standardized parsers. Custom formats like Mermaid and Graphviz are less straightforward, while branching PMRs are penalized due to the need for recursive parsing. We remove one point to BPMN for editability due to the need to update the associated BPMN diagram when making a change to the process. We also account for LLM schema-following capabilities (see \Cref{tab:pmr_main_features}) by adding an additional point to JSON PMRs.

Extensibility grades are based on the inherent limitations of PMRs and effort needed to support additional element types (e.g., it is harder to update the POWL generation logic than the PME JSON schema). 

\begin{table}[t]
\caption{Evaluation summary of PMRs for LLM-based PMo. All grades are from 1 to 5.}
\label{tab:pmr_for_pmo_summary}
\begin{tabularx}{\textwidth}{@{}lRRRRRRR@{}}
\toprule
\textbf{PMR} & 
\textbf{Avg.} & 
\textbf{\begin{tabular}[c]{@{}r@{}}Token\\ compact\end{tabular}} & 
\textbf{\begin{tabular}[c]{@{}r@{}}Expressive\end{tabular}} & 
\textbf{\begin{tabular}[c]{@{}r@{}}Human\\ readable\end{tabular}} & 
\textbf{\begin{tabular}[c]{@{}r@{}}Vizualisable\end{tabular}} & 
\textbf{Usable} & 
\textbf{Extensible} \\
\midrule
BPMN           & 3.33             & 1                          & \textbf{5 \textit{(100\%)}}     & 1                          & \textbf{5}                          & 3                  & \textbf{5}             \\
BPMN process   & 3.50             & 2                          & \textbf{5 \textit{(100\%)}}     & 2                          & 3                                   & 4                  & \textbf{5}             \\
Graphviz       & 3.67             & 4                          & 4 \textit{(89\%)}               & 3                          & \textbf{5}                          & 3                  & 3                      \\
Mermaid        & \textbf{4.00}    & \textbf{5}                 & 4 \textit{(89\%)}               & 4                          & \textbf{5}                          & 3                  & 3                      \\
PME            & 3.20             & 2                          & \textbf{5 \textit{(100\%)}}     & 2                          & 2                                   & \textbf{5}         & 4                      \\
Simplified XML & 3.50             & 3                          & \textbf{5 \textit{(100\%)}}     & 3                          & 2                                   & 3                  & \textbf{5}             \\
POWL code      & 2.83             & \textbf{5}                 & 1 \textit{(71\%)}               & 3                          & 3                                   & 3                  & 2                      \\
BPMN text      & 2.67             & 4                          & 2 \textit{(84\%)}               & \textbf{5}                 & 1                                   & 2                  & 2                      \\
JSON branches  & 3.00             & 4                          & 3 \textit{(87\%) }              & \textbf{5}                 & 1                                   & 3                  & 2                      \\ 
\bottomrule
\end{tabularx}
\end{table}

\subsection{PMR for PMG}

\subsubsection{Element counts.}

\Cref{tab:element_counts} presents the mean element counts of the generated process models for each PMR. On average, LLMs generate significantly smaller models, with roughly eight fewer nodes (tasks, events, or gateways) compared to the ground truth. Gateways are the most affected: the number of Exclusive gateways drops by 50\%, and Parallel gateways by 65\%. This indicates a tendency to simplify the model by retaining a single task sequence.
Despite this general trend, differences between PMRs are notable. Branching PMRs (i.e., JSON branches and BPMN text) lead to substantially higher element counts than the other representations. This suggests that the branching structure can partially mitigate the LLMs' tendency to omit gateways.
Interestingly, Graphviz stands out by producing more tasks and gateways than other graph-based PMRs. This may be attributed to its use of descriptive node names instead of identifiers, reducing generation complexity for the LLM.

\begin{table}[t]
\centering
\caption{Mean element counts of generated PMR models compared to ground truth. Nodes include tasks, events and gateways. Sequence flows are reported separately as they directly depend on the number of nodes and would artificially increase the number of elements.}
\label{tab:element_counts}
\begin{tabularx}{\textwidth}{@{}lRRRRRR@{}}
\toprule
\textbf{PMR} & 
\textbf{Nodes} & 
\textbf{Tasks} & 
\textbf{Events} & 
\begin{tabular}[c]{@{}r@{}}
\textbf{Exclusive} \\
\textbf{gateways}
\end{tabular} & 
\begin{tabular}[c]{@{}r@{}}
\textbf{Parallel} \\
\textbf{gateways}
\end{tabular} & 
\begin{tabular}[c]{@{}r@{}}
\textbf{Sequence} \\
\textbf{flows}
\end{tabular} \\
\midrule
Ground Truth   & 23.18          & 12.53          & 2               & 5.91                        & 2.67                       & 27.67                   \\
\midrule
BPMN           & -10.49         & -4.62          & +0.18           & -4.09                       & -1.98                      & -13.93                  \\
BPMN process   & -9.93          & -4.35          & +0.14           & -3.96                       & -1.69                      & -13.21                  \\
Graphviz       & -7.96          & -2.71          & +0.04           & -3.82                       & \textbf{-1.40}             & -10.29                  \\
Mermaid        & -9.16          & -3.87          & +0.07           & -3.75                       & -1.53                      & -11.80                  \\
PME            & -9.56          & -4.02          & +0.09           & -3.73                       & -1.82                      & -12.69                  \\
Simplified XML & -9.49          & -4.36          & +0.09           & -3.51                       & -1.64                      & -12.38                  \\
POWL code      & -9.36          & -4.08          & \textbf{0.00}   & -3.91                       & -1.67                      & -12.94                  \\
BPMN text      & \textbf{-3.86} & \textbf{-1.53} & \textbf{0.00}   & -0.41                       & -1.85                      & \textbf{-6.01}          \\
JSON branches  & -4.29          & -2.44          & \textbf{0.00}   & \textbf{+0.09}              & -1.87                      & -6.07                   \\
\midrule
Average        & -8.23          & -3.55          & +0.07           & -3.01                       & -1.72                      & -11.04                  \\ 
\bottomrule
\end{tabularx}
\end{table}

\subsubsection{PME similarity.}

\Cref{tab:pme_similarity} reports PME similarity scores obtained for different element types.
Branching PMRs-particularly BPMN text-consistently achieve higher similarity scores than graph-based ones.
Graphviz also shows strong performance for gateway decisions, likely benefiting once again from its use of descriptive identifiers.
POWL code scores are lowered by the high numbers of formatting errors (40\% of generated models are invalid), reflecting the PMR shortcomings with a limited prompting budget.

\begin{table}[t]
\centering
\caption{Mean PME similarity scores of generated PMR models compared to ground truth for each PMR under standard PMG.}
\label{tab:pme_similarity}
\begin{tabularx}{\textwidth}{@{}lRRRRRRR@{}}
\toprule
\textbf{PMR} & \textbf{Overall} & 
\begin{tabular}[c]{@{}r@{}}
\textbf{Tasks} \\
\textbf{overall}
\end{tabular} & 
\begin{tabular}[c]{@{}r@{}}
\textbf{Events} \\
\textbf{overall}
\end{tabular} & 
\begin{tabular}[c]{@{}r@{}}
\textbf{Gateways} \\
\textbf{overall}
\end{tabular} & 
\begin{tabular}[c]{@{}r@{}}
\textbf{Gateway} \\
\textbf{decisions}
\end{tabular} & 
\begin{tabular}[c]{@{}r@{}}
\textbf{Gateway} \\
\textbf{types}
\end{tabular} & 
\begin{tabular}[c]{@{}r@{}}
\textbf{Sequence} \\
\textbf{flows}
\end{tabular} \\ 
\midrule
BPMN           & 0.43             & 0.48                   & 0.55                    & 0.38                      & 0.16                       & 0.45                   & 0.36                    \\
BPMN process   & 0.44             & 0.48                   & 0.55                    & 0.4                       & 0.15                       & 0.5                    & 0.38                    \\
Graphviz       & 0.47             & 0.5                    & 0.74                    & 0.49                      & \textbf{0.6}               & 0.54                   & 0.38                    \\
Mermaid        & 0.48             & 0.49                   & 0.74                    & 0.45                      & 0.18                       & 0.54                   & 0.42                    \\
PME            & 0.46             & 0.5                    & 0.68                    & 0.45                      & 0.12                       & 0.54                   & 0.37                    \\
Simplified XML & 0.45             & 0.48                   & 0.57                    & 0.46                      & 0.07                       & 0.58                   & 0.4                     \\
POWL code      & 0.27             & 0.29                   & 0.41                    & 0.21                      & 0.36                       & 0.23                   & 0.22                    \\
BPMN text      & \textbf{0.54}    & \textbf{0.52}          & \textbf{0.75}           & 0.58                      & 0.13                       & 0.69                   & 0.49                    \\
JSON branches  & 0.53             & 0.51                   & 0.58                    & \textbf{0.61}             & 0.19                       & \textbf{0.7}           & \textbf{0.5}            \\ 
\bottomrule
\end{tabularx}
\end{table}

\section{Discussion and Conclusion}\label{sec:conclusion}
This study offers a comprehensive analysis of PMRs in the context of PMo with LLMs, with a particular focus on the PMG task.

To support this investigation, we assemble the largest gold-standard dataset to date for PMo (55 pairs) coupled with various types of PMRs, enabling both qualitative and quantitative comparisons across \pmrnumber PMRs drawn from the literature. Our analysis, guided by a set of functional requirements, identifies Mermaid as the most suitable PMR for PMo with LLMs, notably due to its token compactness and visualization capabilities.

We further examine the use of PMRs as generation targets for PMG. After generation based on standard prompting, we measure performance using process element counts and PME similarity. Results reveal a consistent tendency of LLMs to undergenerate process elements—particularly gateways—leading to simplified process models. However, PMRs that support explicit branching structures, such as BPMN text and JSON branches, mitigate this behavior. BPMN text, in particular, achieves the strongest similarity with ground truth elements for both raw numbers and semantic content.

Overall, our findings highlight the critical role of PMR design in enabling effective LLM-based process modeling and generation. While Mermaid appears to be the most versatile PMR, BPMN text yields better results for PMG. This suggests that using different PMRs at specific stages of the PMo pipeline may lead to more effective outcomes.

Several limitations must be acknowledged and provide avenues for future research.

First, our evaluation of PMRs for PMo with LLMs (\Cref{tab:pmr_for_pmo_summary}), while incorporating quantitative metrics, remains primarily subjective, as it is based on the assessment of the authors. A more rigorous approach, such as combining the grades from multiple experts, would enhance the validity of our findings.

Second, the structure of PMRs is inherently flexible, and our evaluation relies on standardized representations (e.g., consistent identifier formats). Alternative formatting choices may influence LLM behavior and lead to different results. Moreover, several promising PMRs, such as BPMN Sketch~\cite{ivanchikj_live_2022} or JSON-Nets~\cite{forell_modeling_2024}, were not included in this study and represent valuable directions for future exploration.

Third, another limitation is that PME similarity, although it captures semantic content,  remains highly sensitive to the number of generated elements. While most generated elements match the ground truth, omissions significantly reduce the overall score. For this reason, we also report element counts. Moreover, it enables us to better understand the under-generation problem of LLMs.

Fourth, we do not manually evaluate the LLM-generated models, which could provide better insights of their quality and practical usability.

Finally, our PMG experiments, like most existing studies, focus on a reduced subset of process elements. As such, they do not capture the full expressive range of real-world models. 
Future research should investigate LLMs’ ability to generate richer models that include more advanced elements such as swimlanes or data objects.

\subsubsection*{\ackname} This study was funded by MITACS grant number IT32670.

\subsubsection*{Declarations of Interest.}
The authors have no competing interests to declare that are relevant to the content of this article.

\bibliographystyle{splncs04}
\bibliography{bib/all_tidy}

\end{document}